\documentclass[runningheads]{llncs}

\usepackage[utf8]{inputenc}
\usepackage{graphicx}
\usepackage{amssymb,amsmath,mathtools}
\usepackage{enumitem}
\usepackage[pdftex, pdfborder={0 0 0}, hyperfootnotes=false]{hyperref}
\usepackage{url}
\usepackage[textsize=scriptsize, disable]{todonotes}
\usepackage{pgfplots}
\usepackage{tikz}
\usepackage{csquotes}
\usepackage{siunitx}
\usepackage{marvosym}
\usepackage{multirow}
\usepackage{caption,subcaption}

\pgfplotsset{compat=1.18}

\newcommand{\argmin}{\ensuremath{\operatorname{arg\,min}}}

\newcommand{\refsec}[1]{Section~\ref{#1}}
\newcommand{\reffig}[1]{Fig.~\ref{#1}}
\newcommand{\reftab}[1]{Table~\ref{#1}}
\newcommand{\refeqn}[1]{Eq.~\refeq{#1}}
\newcommand{\refeqc}[1]{(\refeq{#1})}

\begin{document}
\title{Accelerating $k$-Means Clustering\\ with Cover Trees%
\thanks{Part of the work on this paper has been supported by Deutsche Forschungsgemeinschaft (DFG),
project number 124020371,
within the Collaborative Research Center SFB 876
``Providing Information by Resource-Constrained Analysis'', project A2.
}}
\author{
    Andreas~Lang\Letter\orcidID{0000-0003-3212-5548}
    \and
    Erich~Schubert\orcidID{0000-0001-9143-4880}
}
\titlerunning{Accelerating $k$-Means Clustering with Cover Trees}
\authorrunning{A. Lang and E. Schubert}
\institute{TU Dortmund University, Dortmund, Germany
    \email{\{andreas.lang,erich.schubert\}@tu-dortmund.de}
\\
\textcolor{red}{
\textbf{Preprint}. Please consult the final version instead:
\\
\url{https://doi.org/10.1007/978-3-031-46994-7_13}
\vspace{-7mm}
}
}
\date{}%
\maketitle
\begin{abstract}
$k$-means clustering is a popular algorithm that partitions data into $k$~clusters.
There are many improvements to accelerate the standard algorithm.
Most current research employs upper and lower bounds on point-to-cluster distances
and the triangle inequality to reduce the number of distance computations,
with only arrays as underlying data structures. These approaches cannot exploit
that nearby points are likely assigned to the same cluster.
We propose a new $k$-means algorithm based on the cover tree index, that has
relatively low overhead and performs well, for a wider parameter range, than
previous approaches based on the k-d tree.
By combining this with upper and lower bounds, as in state-of-the-art approaches,
we obtain a hybrid algorithm that
combines the benefits of tree aggregation and bounds-based filtering.

\end{abstract}

\section{Introduction}
One of the most popular clustering algorithms is \mbox{$k$-means},
often with the standard algorithm taught in textbooks (commonly attributed to Lloyd \cite{DBLP:journals/tit/Lloyd82}, but described before by,
e.g., Steinhaus \cite{Ste56}). %
In $k$-means, the data is approximated using $k$ centers, which are the arithmetic mean of the partitions,
and the goal is to minimize the sum of squared deviations of all samples and their nearest centroids.
Finding the true optimum is NP-hard~\cite{DBLP:conf/walcom/MahajanNV09}, and hence we need %
heuristics such as the standard algorithm.
The popularity of $k$-means and an ever-increasing amount of data led to many improvements to the standard algorithm.
Most common improvements replicate the convergence of the standard heuristic exactly,
and are hence sometimes called ``exact'' $k$-means. %
The major part of its runtime is the distance calculations between samples and cluster centers in each iteration.
One way to accelerate $k$-means is to approximate the data, e.g., by sampling~\cite{DBLP:conf/kdd/BradleyFR98} or aggregation,
which is used in mini-batch $k$-means~\cite{DBLP:conf/www/Sculley10},
BICO~\cite{DBLP:conf/esa/FichtenbergerGSSS13}, and BETULA~\cite{DBLP:journals/is/LangS22}, among others.
The expected values of the results are very similar to the standard algorithm, which is not surprising because the means used in $k$-means are statistical summaries, too. 
For ``exact'' $k$-means, without lossy data aggregation, approaches primarily fall into two categories:
(1)~k-d trees have been used to accelerate k-means~\cite{DBLP:conf/kdd/PellegM99,DBLP:journals/pami/KanungoMNPSW02}
by assigning subsets of the data to clusters at once, using the distances of the centers to the bounding boxes of the tree nodes,
and (2)~a large family of approaches which use the triangle inequality to omit unnecessary distance computations,
exploiting that many points do not change their cluster after the first few iterations.
Philips~\cite{DBLP:conf/alenex/Phillips02} used the pairwise distances of the centers to identify unnecessary
computations, and Elkan~\cite{DBLP:conf/icml/Elkan03}
additionally uses upper and lower bounds for the distances between points and cluster centers.
Hamerly~\cite{DBLP:conf/sdm/Hamerly10} merged the lower bounds to the far cluster centers into a single bound to conserve memory, at the cost of additional distance computations due to looser bounds.
State-of-the-art $k$-means algorithms, like Exponion~\cite{DBLP:conf/icml/NewlingF16}
and Shallot~\cite{DBLP:conf/ida/Borgelt20}, additionally use \mbox{(hyper-)}balls around the
closest centers to further reduce computations.
None of these methods is always best, but it depends on data dimensionality, data size,
the number of clusters, initialization, and the data distribution. As all of these methods compute all
distances between points and initial centers in the first round (to obtain the initial bounds), the first iteration is at least
as expensive as in the standard algorithm, but it is in the later iterations %
where these improvements help.
Recent proposals, for example, also take the distances to the previous center locations into account~\cite{DBLP:conf/sisap/YuCC20},
or transfer such acceleration techniques to spherical $k$-means~\cite{DBLP:conf/sisap/SchubertLF21} for text clustering by
using a similar triangle inequality for cosine similarity~\cite{DBLP:conf/sisap/Schubert21}.

In this paper, we combine the ideas from both of these research directions:
we will use an exact data index (a trivial extension of the cover tree~\cite{DBLP:conf/icml/BeygelzimerKL06}),
combined with a pruning strategy that uses the triangle inequality, to accelerate the standard $k$-means algorithm
by assigning entire subsets of the data at once.
We then use our approach to calculate the upper and lower bounds that are used in Hamerly's~\cite{DBLP:conf/sdm/Hamerly10} algorithm and its derivatives. Which allows us to switch strategies in later iterations
when the clusters have become stable and these bounds become effective.
We choose the cover tree as it aggregates the data into a hierarchy of ball covers,
which allows the direct use of the triangle inequality.
This differs from previous approaches using the k-d tree, which used the minimum distance to the
bounding boxes~\cite{DBLP:conf/kdd/PellegM99},
respectively hyperplanes implied by these bounding boxes~\cite{DBLP:journals/pami/KanungoMNPSW02};
both of these methods need bounding boxes not used by the original k-d tree.
We argue that the ball covers of the cover tree produce more suitable bounds than the bounding boxes used in the k-d tree approaches,
and that metric pruning should be superior to the geometric pruning of existing approaches.
Furthermore, we hope to achieve improved performance by the way the cover tree controls the data expansion
rate (and hence the radius of the nodes). It allows a wider fan-out,
whereas the \mbox{k-d} tree is a strict binary tree that will have a higher depth, and more nodes. 
Last, but not least, a node in the (extended) cover tree is a more compact data structure than the bounding boxes used
by the k-d tree approaches by approximately a factor of two: a ball is represented by a center and a radius,
whereas the bounding boxes are represented by the midpoint and width in each axis (or alternatively, by a minimum and maximum,
but the former is more suitable for $k$-means). Hence, we expect the new approach to need less memory.

\section{Foundations}
In this section, we explain the foundations of the cover tree and $k$-means, using a notation inspired by previous work \cite{DBLP:conf/icml/NewlingF16} suitable for both.

\subsection{Standard \boldmath$k$-Means Algorithm}
The standard $k$-means algorithm is a heuristic to partition the points $s \in X$ into $k$ clusters $\{C_1, ..., C_k\}$.
We start with the initial cluster centers $c_1, ..., c_k$ (e.g., sampled from the data)
and alternate the following two optimization steps:
first, all samples~$s$ are assigned to their nearest cluster center by Eq.~\refeqc{eq:ass},
where $a(s)$ denotes the cluster index, and then
the cluster centers $c_i$ are updated by Eq.~\refeqc{eq:upd}:
\begin{align}
    a(s) &\leftarrow \argmin_{i \in 1,..,k} d(s,c_i), s \in X \label{eq:ass}\\
    c_i & \leftarrow \frac{\sum_{s|a(s)=i}s}{\vert \{s \mid a(s) = i\} \vert} \quad i \in 1,...,k \label{eq:upd}.
\end{align}
When no cluster assignment changes, the algorithm stops.%
\footnote{It is also possible to stop early centers' movement is below some threshold.}
While this heuristic does frequently not find the global optimum, this scheme converges to a local fix point %
when no assignment changes.
The selection of the initial cluster centers not only influences the run time of the algorithm but also the quality of the final clustering.
Here, the $k$-means++~\cite{DBLP:conf/soda/ArthurV07} initialization has become the most popular choice,
as it already provides a (probabilistic) quality guarantee, based on sampling centers from the data
proportionally to their expected contribution to the reduction of variance.
In this article, we will not further study initialization.

\subsection{Triangle Inequality in \boldmath$k$-Means}
\label{sec:kmbounds}

Many accelerated $k$-means algorithms use the triangle inequality to reduce the number of distance calculations,
in particular in Eq.~\refeqc{eq:ass} instead of computing the distances between all points and all clusters, as introduced by Phillips \cite{DBLP:conf/alenex/Phillips02}.
Given sample~$s$ and cluster center~$c_i$ for which the distance $d(s,c_i)$ is known,
and some other $c_j$ with an unknown distance, we can use the triangle inequality
\begin{alignat}{2}
    d(c_i,c_j) &\leq d(s,c_i) &{}+{}& d(s,c_j)\label{eq:tri-upper}\\
\Rightarrow\quad
    d(s,c_j) &\geq d(c_i,c_j) &{}-{}& d(s,c_i)\label{eq:tri-lower}
\end{alignat}
to get a lower bound on the distance $d(s,c_j)$, and therefore may be able to avoid computing $d(s,c_j)$.
In particular, Eq.~\refeqc{eq:tri-lower} yields the implication \cite{DBLP:conf/alenex/Phillips02}:
\begin{align}
    &d(c_i,c_j) \geq 2d(s,c_i) 
    \quad \Rightarrow \quad
    d(s,c_j) \geq d(s,c_i)
    \;.
    \label{eq:innerkm}
\end{align}

We can exclude centers from consideration that are far away from the current closest center with this inexpensive filter.
By computing all $d(c_i,c_j)$ once at the beginning of each iteration,
this is relatively cheap as long as $k$ is not too big.

Another application for the triangle inequality in $k$-means are per-sample upper bounds on the distance to the assigned cluster and lower bounds on the distances to the other cluster means:
\begin{align*}
    u_s \geq d(s, c_{a(s)}) \qquad \text{ and } \qquad
    l_s \leq d(s, c_{j}) \qquad \forall j \neq a(s).
\end{align*}
Then $u_s\leq l_s$ implies that $a(s)$ is still the closest cluster by $d(s, c_{a(s)}) \leq u_s \leq l_s \leq d(s, c_{j})$. 
These bounds were introduced by Elkan~\cite{DBLP:conf/icml/Elkan03} and later simplified by Hamerly~\cite{DBLP:conf/sdm/Hamerly10} to use only a single lower bound for all other clusters.
We increase the upper bound and decrease the lower bound based on the movement of the centers to guarantee their correctness.
This becomes more effective once the cluster centers move only very slightly: while cluster centers still move much, we may see $u_s>l_s$ often, and then have to compute the true distances often.
For Hamerly's version, a single cluster that moves substantially is enough to require many recomputations,
Elkan's version computes fewer distances, but has to update $N\cdot k$ bounds each iteration.
The current state-of-the-art algorithms Exponion~\cite{DBLP:conf/icml/NewlingF16} and Shallot~\cite{DBLP:conf/ida/Borgelt20} also use the same bounds as Hamerly.

When the distances of a sample $s$ to all cluster means $c_i$ are calculated, the upper bound $u_s$ is set to the distance to the assigned cluster $a(s)$ and the lower bound $l_s$ to the minimum distance to any of the other clusters $c_j$ for $j\neq a(s)$:
\begin{align*}
    u_s &\leftarrow d(s,c_{a(s)}) & \qquad
    l_s &\leftarrow \min_{j \neq {a(s)}} d(s,c_j).
\end{align*}
In each iteration, when updating the means, we also have to update the bounds.
For each mean $c_i$, we compute how far it moved from its previous location $c'_i$.
To retain correct bounds, we have to add the distance moved by the nearest mean to the upper bound,
and subtract the maximum distance moved by any other cluster mean from the lower bound:
\begin{align*}
    u_s &\leftarrow u_s + d(c'_{a(s)}, c_{a(s)})&\qquad
    l_s &\leftarrow l_s - \max_{j \neq {a(s)}} d(c'_j, c_j). 
\end{align*}

The various algorithm variants proposed often include additional pruning rules using the triangle
inequality, but a thorough discussion is beyond the scope of this paper, see \cite{DBLP:conf/icml/NewlingF16,DBLP:conf/ida/Borgelt20} for an overview of recent algorithms.

\subsection{Cover Tree}
\label{sec:covTree}
The cover tree \cite{DBLP:conf/icml/BeygelzimerKL06} is a tree-based index structure with linear memory designed to accelerate nearest neighbor and radius search.
The key idea of the cover tree is to cover the data with balls of a radius that decreases as we move down the tree;
in its theoretical formulation it has an infinite number of levels with radius~$2^i$;
but in practice, there exists a top level where all data is in a single ball and a bottom level where
all distinct points have been separated. Levels in-between can be omitted if no changes to the tree structure happen.
If the dataset has a finite expansion rate
(i.e., the amount of data grows by at most a constant if we double the radius), this index can provide
interesting theoretical guarantees for nearest neighbor search.
In practice, the cover tree often performs quite well because of its small overhead and as it is inexpensive to build.
The ball covers of the cover tree %
are restricted by the maximum radius
in each level, and the tree structure: a sample~$s$ that was contained in a node~$x$
must remain in a child of~$x$ in the next level, and must not move to another ball even if that were closer.

All cover trees obey invariants for their covers $N_i$ at each level $i\in\mathbb{Z}$:
\begin{enumerate}
    \item (nesting) $N_i \subset N_{i-1}$
    \item (cover) $\forall q \in N_{i-1} \exists p \in N_i ~{d(p,q) \leq 2^i}$ and exactly one $p$ is the parent of $q$
    \item (separation) $\forall p,q \in N_i, d(p,q) \ge 2^i$.
\end{enumerate}
The $p \in N_i$ function as routing objects for searching the tree, and at each level balls of radius~$2^i$ around these objects cover the entire dataset.
Instead of storing the $N_i$ at infinitely many levels, we only store levels that differ
from the previous. To make the tree navigable from the root, we store for every $p\in N_i$
all $q \in N_{i-1}$ for which $p$ is the parent, and their distance $d(p,q)$; except $p$ itself which always is its own child at distance~0.
An interesting side effect of optimization is that since $p$ then is also a routing object in the next level,
we can reuse any distance to~$p$ that we already computed in all subsequent levels.
Two routing objects at the same level have at least the distance $d(p,q) > 2^i$,
while descendants are within the radius $d(p,q) \leq 2^i$ to the routing object.
While the factor $2^i$ was used for theoretical results,
smaller scaling factors of 1.2--1.3 are typically faster in practice. 
The scaling factor allows controlling the trade-off between fan-out (width) and depth of the tree.
We use the scaling factor of~$1.2$ in our experiments.

\hyphenation{Bey-gel-zi-mer}%
The construction of the cover tree follows the original greedy approach of Beygelzimer et al.~\cite{DBLP:conf/icml/BeygelzimerKL06}.
We extend it with a simple bottom-up aggregation afterward to store the sum $S_x = \sum_{p \in x}p$ and the number of samples $w_x$ in each node~$x$.
While this increases the memory consumption of the cover tree noticeably
(previously, it would store only object references and the distances to the parent),
we still have to store only one vector for each node. Not two vectors for the bounding box as in the earlier k-d tree approaches.
Because of the higher fan-out, we also have fewer nodes than the k-d tree. 
Furthermore, we can define a minimum node size, at which we stop building the tree and instead
store all remaining points directly with a cover radius of 0.
For efficiency, we store all such singleton nodes ($|x|=1$) more compactly, and omit storing the
trivial aggregated values $S_x=x$ and $w_x=1$, and the radius $r=0$.

\subsection{Bounds within a Cover Tree}
\label{sec:cvbounds}

To assign an entire cover tree node that represents a subset of the dataset $x \subseteq X$ to a cluster, all samples $q_x\in x$ have to be closest to the same center.
We can bound distances using the node radius $r_x =  \max_{q_x\in x} d(p_x,q_x)$:
\begin{alignat}{3}
    d(p_x,c_i)-r_x &\leq  &~d(q_x,c_i) &\leq d(p_x,c_i) + r_x \label{eq:x-prc}
\;.\end{alignat}
The triangle inequality also yields bounds on the distance of the routing object $p_y$ %
to a cluster center $c_i$, given the distance from the parent node to the cluster:
\begin{alignat}{3}
    d(p_x,c_i) - d(p_x,p_y) &\leq d(p_y,c_i) %
    \leq d(p_x,c_i) + d(p_x,p_y) \label{eq:x-ppc}
\;.\end{alignat}
Combining these yields upper and lower bounds for all samples in a child node,
which we can use to prune candidate centers in $k$-means when $\forall q_y \in y$:
\begin{align}
    d(p_x,c_j) - d(p_x,p_y) - r_y 
    & \leq d(q_y,c_j) %
     \leq d(p_x,c_j) + d(p_x,p_y) + r_y
    \label{eq:y-prc}
\;.\end{align}

\section{Cover Tree \boldmath$k$-Means}
We present a novel algorithm that uses a variation of the cover tree index to filter candidate centers.
Using the triangle inequality we calculate bounds, see \refsec{sec:cvbounds}, to rule out cluster centers and hence reduce the number of distance computations in the assignment phase (i.e., in \refeqn{eq:ass}) of $k$-means, using a similar idea as the k-d tree-based approaches \cite{DBLP:conf/kdd/PellegM99,DBLP:journals/pami/KanungoMNPSW02}.
However, the new approach uses fewer and smaller nodes for the tree, and as routing objects are reused in the cover tree, needs fewer distance computations.
Tree nodes (representing subsets of the data) can be assigned at once, which is particularly beneficial with near-duplicate points.
We then further combine this approach with current state-of-the-art stored-bounds $k$-means algorithms.

\subsection{Calculating Distances}
\label{sec:inner}
During initialization, or when reassigning points, we need the distance from some routing object $p_x$ of cover tree node~$x$ to the centers~$c_i, c_j$ of a set of candidate clusters $C_i, C_j \in A_x$. %
For this, we adapt the common lower bound in \refeqn{eq:innerkm}.
But unlike existing algorithms that handle only single samples, we need to ensure that the bounds are true for every sample $q_x \in x$ represented by the node.
To avoid computing all distances and to benefit from the tree structure, we use \refeqn{eq:x-prc} and the maximum distance $r_x$ of points in the cover tree node. %
This allows deciding whether we have to calculate the distance $d(q_x, c_j)$ based on the already calculated distance $d(q_x, c_i)$ and the inter-cluster distances $d(c_i,c_j)$:
\begin{align}
    &d(c_i,c_j)  \geq 2d(p_x, c_i) + 2r_x \geq 2d(q_x, c_i) \label{eq:comp_means}\\
    & \quad\Rightarrow d(q_x, c_j) \geq d(q_x, c_i) \Rightarrow C_i \notin A_x \nonumber
\;.\end{align}
The inter-cluster distances are computed and stored at the beginning of each $k$-means iteration,
and used many times, as in previous work.

\subsection{Assigning Nodes}
With all relevant bounds for the routing object $p_x$ computed, node $x$ can be assigned to a cluster $a(x) = C_1$ if $\forall q_x \in x$, the distance to the nearest cluster center $c_1$ is smaller than the distance to the second-nearest cluster $c_2$.
To decide this, we use the triangle inequality applied to a node, see \refeqn{eq:x-prc}:
\begin{align}
     d(q_x,c_1) &\leq d(p_x,c_1)+r_x %
    \leq d(p_x,c_2)-r_x \leq d(q_x,c_2). \label{eq:assx}
\end{align}
Especially for the higher levels of the tree, this inequality often does not hold
because nodes have a large radius~$r_x$. Nevertheless, we may be able to eliminate candidates
$C_i \in A_x$ for subsequent levels.
To rule out that a cluster $C_i$ is the nearest to any sample $q_x \in x$, the distance from its centroid $c_i$ to the routing object $p_x \in x$ has to be larger than the upper bound on the distance to its nearest cluster center,
which is a generalization of \refeqn{eq:assx}:
\begin{align}
    d(q_x,c_1)  &\leq d(p_x,c_1)+r_x %
    \leq d(p_x,c_i)-r_x \leq d(q_x,c_i). %
    \label{eq:filterCx}
\end{align}
As we narrow down the set of candidate cluster centers $A_x$, eventually only a single center remains, and the entire
subtree can be assigned to the same cluster.

If we reassign the node and all its contained points to another cluster, we remove  $S_y$ and $w_y$ for all previously assigned nodes $y \subseteq x$ from their old clusters and add $S_x$ and $w_x$ to the new cluster instead.
By using the aggregates stored in the tree, this reassignment becomes more efficient.
If we cannot assign a node, we process all child nodes recursively.

\subsection{Recursion into Child Nodes}
When moving to a child node $y\subset x$ in the tree, we can exploit that we store the distances to the parent in the cover tree, and that we know the radius of~$y$.
The inequality \refeqn{eq:y-prc} combines the known distances to the parent routing object $d(p_x,c_i)$,
the distance between parent and child routing objects $d(p_x,p_y)$,
and the reduced radius of the child $r_y$.
We now can assign the entire child to the parents' nearest cluster $C_1$ if $\forall q_y \in y$:
\begin{align}
    d(q_y,c_1) &\leq %
     d(p_x,c_1) {+} d(p_x,p_y) {+} r_y %
     \leq d(p_x,c_2){-} d(p_x,p_y) {-}r_y 
     \leq d(q_y,c_2). %
    \label{eq:assy}
\end{align}
Observe that for $p_x=p_y$ we obtain \refeqn{eq:assx} with the reduced radius of the next level,
but this bound is only tighter if $d(p_x,p_y)\leq r_x-r_y$.
Since $d(p_x,p_y)>r_y$ for $p_x\neq p_y$ and the radius $r_y$ typically only reduces by the chosen factor~$1.2$,
it frequently will not be sufficient. But as the tree can skip some levels, it occasionally holds and allows skipping some computations.
If the above inequality does not hold, we have to recompute at least some distances to assign $y$.
The distance $d(p_y,c_1)$ to the nearest cluster center $c_1$ of the parent node is most beneficial because it can be used to tighten the inequality in \refeqn{eq:assy} by eliminating the distance to the parent node on the left-hand side of the equation and allows us to assign $y$ to $C_1$
 if $\forall q_y \in y$:
\begin{align}
    d(q_y,c_1) &\leq %
    d(p_y,c_1) + r_y %
    \leq d(p_x,c_2)- d(p_x,p_y) -r_y
    \leq d(q_y,c_2). %
    \label{eq:assy_tight}
\end{align}
The same tightening is applied to \refeqn{eq:filterCx} to prune candidate clusters before also recomputing
the distances $d(q_y, c_i)$ when moving to the child node.
Clusters that do not satisfy the following inequality can be excluded:
\begin{align}
    d(q_y,c_1) & \leq %
    d(p_y,c_1) + r_y %
    \leq d(p_x,c_i)- d(p_x,p_y) -r_y
    \leq d(q_y,c_i). %
    \label{eq:filterCy}
\end{align}

After pruning cluster centers with \refeqn{eq:filterCy}, the same bounds and calculations are applied recursively
to each of the child nodes not yet assigned a cluster.

\subsection{Hybrid Cover Tree \boldmath$k$-Means}\label{sec:hybrid}
When analyzing the number of distance computations performed per iteration by
current state-of-the-art algorithms, like Exponion~\cite{DBLP:conf/icml/NewlingF16}
and Shallot~\cite{DBLP:conf/ida/Borgelt20},
we observe that they correlate with how much the cluster centers move, for the reason explained before.
They decrease drastically over the first few iterations as the centers stabilize.
Our approach, on the other hand, is only slightly influenced by how far means move,
and can already avoid distance computations in the first iteration by reducing the number of candidate cluster
centers when traversing the tree.
Using the stored aggregates in the tree, we can assign entire subtrees to the same cluster, but the pure tree-based approach benefits little from centers becoming stable after a few iterations.
To utilize the best of both worlds, we propose a hybrid algorithm that uses the cover tree only
for the first few iterations and then switches to the state-of-the-art
Shallot $k$-means algorithm~\cite{DBLP:conf/ida/Borgelt20}.
Any other algorithm based on Hamerly's~\cite{DBLP:conf/sdm/Hamerly10} bounds could be used instead,
as we can efficiently obtain upper and lower bounds from our cover tree, which gives these algorithms
an efficient start. Obtaining all $k\cdot N$ bounds for Elkan's algorithm would be more effort.
Our approach is not equivalent to initializing with cluster centers obtained with the cover tree,
but we prune distance computations when computing the bounds using the tree by filtering candidate means.

For single points, this is trivial (but we still save, as we may have a reduced set of candidates for the second-nearest center). %
If we assign an entire node $x$ to a cluster $C_1$, we do not know the exact distance to the nearest two clusters,
but we obtain upper and lower bounds by our inequalities above:
\begin{align}
    u_{q \in x} &= d(p_x,c_1) + r_x \\
    l_{q \in x} &= d(p_x,c_2) - r_x. 
\end{align}
When assigning a subtree, we can also obtain bounds for child nodes $y \subset x$
without additional distance computations:
\begin{align}
    u_{q \in y} &= d(p_x,c_1) + d(p_x,p_y) + r_y \\
    l_{q \in y} &= d(p_x,c_2) - d(p_x,p_y) - r_y. 
\end{align}
While computing $d(p_y,c_1)$ for each child node $y$ would yield tighter bounds,
we can simply leave this to the next iteration of the subsequent $k$-means algorithm.
Depeding on the center movements, these will need to be refined anyway, and do not need to be tight.
For initializing the Shallot algorithm, we also need to give it the identity of the second-nearest cluster;
more precisely, the cluster for which the lower bound was obtained.
While it is not guaranteed that the second-nearest cluster to the routing object in the tree
is the second-nearest cluster for all points in the node,
the Shallot algorithm only requires the bounds to hold.
In the regular Shallot algorithm, it can happen that the assumed second-nearest cluster changes unnoticed,
this does not affect the correctness of the algorithm.

In summary, when we transition to a stored-bounds algorithm, we can use the bounds used by the Cover-means algorithm to initialize the stored-bounds. These bounds will be less tight, but also much less expensive to compute, as computing the initial bounds is a bottleneck of all stored-bounds approaches.

\section{Evaluation}

\textit{Algorithms:}
In our experiments, we compare our new cover-tree-based approaches to the \emph{Standard} $k$-means algorithm, as well as state-of-the-art improvements.
The $k$-d tree filtering variant of \emph{Kanungo} et al.~\cite{DBLP:journals/pami/KanungoMNPSW02}
represents current tree-based methods.
From the stored-bounds family, we include the popular and easy-to-understand
\emph{Hamerly's}~\cite{DBLP:conf/sdm/Hamerly10} and \emph{Elkan's}~\cite{DBLP:conf/icml/Elkan03} $k$-means algorithm,
and the state-of-the-art algorithms \emph{Exponion}~\cite{DBLP:conf/icml/NewlingF16} and \emph{Shallot}~\cite{DBLP:conf/ida/Borgelt20}.
We include two new approaches in our evaluation. 
First, the base variant, \emph{Cover-means}, uses the cover tree for accelerating $k$-means.
Second, a hybrid of our cover tree variant with Shallot (as explained in Section~\ref{sec:hybrid})
denoted as \emph{Hybrid}.

\textit{Parameterization:}
The cover tree brings some additional hyperparameters for the algorithms.
We decided not to tune these parameters for specific datasets, but instead we identified a set of default values
that will usually work well, and do not vary them in the experiments.
Much of these are set to keep the cover tree small, as to not introduce overhead for constructing the tree.
The cover tree is built with a scaling factor of $1.2$; while larger values may further accelerate clustering,
the increased fan-out comes with an increased construction time, in particular for uniform data regions.
To further keep the tree small, we stop splitting at a minimum node size of $100$ samples.
Smaller leaf nodes would allow more pruning, but also increase the construction time.
For the hybrid approach, we switch strategies after $7$ iterations. This value has likely the most potential
for further turning for particular datasets; in particular, for easy synthetic datasets, it may be beneficial to switch later.
On the other hand, on too easy datasets or with lucky initialization, $k$-means may already have converged before.
We include the construction time of the cover tree, except for the experiment listing time per iteration and the parameter sweep experiment.
We evaluated the same 10 random initializations generated by $k$-means++~\cite{DBLP:conf/soda/ArthurV07} for each algorithm,
and usually study medium to large $k=10\ldots 1000$.
\begin{table}[tb]
    \caption{Overview of the datasets used in the experiments}%
    \label{tab:datasets}%
    \centering%
    \setlength{\tabcolsep}{3pt}
    \begin{tabular}{lrcl}
    Name & N &D & domain \\
    \hline
    ALOI & 110250 & 27, 64 & color histograms \\
    MNIST & 70000 & 10, ..., 50 & autoencoder \\
    CovType & 581012 & 54 & remote sensing \\
    Istanbul & 346463& 2 & tweet locations \\
    Traffic & 6.2M& 2 & accident locations \\
    KDD04 & 145751 & 74 & biology 
    \end{tabular}
\end{table}

\textit{Implementation:}
To make the benchmarks more reliable, we keep implementation differences to a minimum and implement all algorithms
in a shared codebase, as recommended by Kriegel et al.~\cite{DBLP:journals/kais/KriegelSZ17}.
We implemented the algorithms in Java using the ELKI framework~\cite{DBLP:conf/sisap/Schubert22}
because it already contains optimized and tested versions of the comparison algorithms.
All algorithms are run single-core on an exclusively used AMD EPYC~7302 CPU to reduce the confounding factors.

\textit{Datasets:}
Because the performance of the $k$-means algorithms strongly depends on the dataset, we evaluate our approach on various real-world datasets, \reftab{tab:datasets}. 
For the MNIST dataset which we reduced in dimensionality with an autoencoder, we chose 10, 20, 30, 40 and 50 dimensions.
The ALOI data is available in multiple dimensionalities~\cite{ALOIMultiView10}, of which we selected 27 and 64 dimensions.
The Istanbul and Traffic datasets contain the coordinates of Tweets respectively traffic accidents, and hence are low-dimensional. 

\begin{figure}[tb!]\centering
    \includegraphics[width=.7\linewidth]{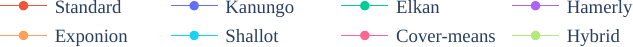}\\
    \begin{subfigure}{.48\linewidth}\centering
        \includegraphics[width=\linewidth]{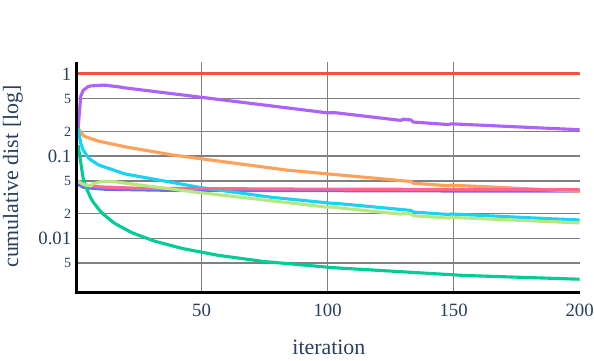}
        \caption{Distance calculations}
        \label{fig:iter_dist}
    \end{subfigure}
    \hfill
    \begin{subfigure}{.48\linewidth}\centering
        \includegraphics[width=\linewidth]{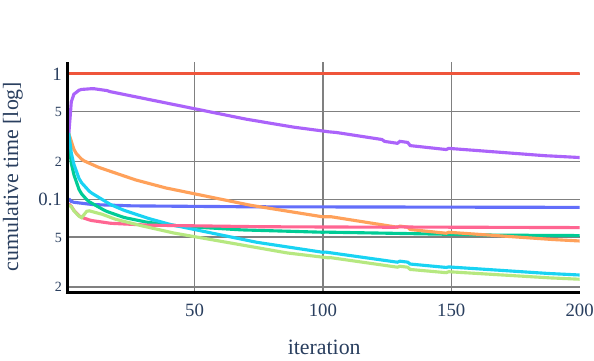}
        \caption{Time}
        \label{fig:iter_time}
    \end{subfigure}
    \caption{Commulative evaluation in relation to the Standard algorithm vs.{} iterations on the ALOI 64D dataset for $k$ = 400}
    \label{fig:iter}
\end{figure}

\textit{Experimental Results:}
In \reffig{fig:iter}, we study the cumulative number of distance computations and the cumulative time on the ALOI~64D dataset for $k{=}400$ over the iterations (i.e., convergence) of the algorithms.
The construction of the tree is not included here. %
Both measurements are normalized by the full Standard algorithm to improve readability,
and can be interpreted as relative savings.

For the number of distance calculations in \reffig{fig:iter_dist}, the algorithms can be categorized into three groups:
The Standard algorithm obviously does not skip any distance calculations.
The tree-based algorithm of Kanungo and the cover tree, only need a fixed fraction of that (5\%-10\% on this data).
When the cluster centers move only a little after the first few iterations, they exhibit a constant performance.
The third group are the stored-bound-based algorithms, which exhibit a decreasing number of distance computations.
The performance of Elkan's algorithm is noticeably better than the others on this metric,
Hamerly's performs worst, while Exponion and Shallot improve over Hamerly's as expected.
Our new Hybrid method combines the early savings of the tree methods with the good late performance
of Shallot (here, switching to Shallot later would likely be better).

\reffig{fig:iter_time} shows the related run time.
The Standard algorithm is again the baseline, and we have similar groups of behavior, but some interesting differences.
The most significant difference is that Elkan's algorithm now is considerably worse.
While Elkan's can save many distance computations, it has to maintain many more bounds, which introduces a constant overhead
per iteration. Exponion and Shallot can save distance computations at little constant overhead, and hence improve over
Hamerly's significantly.
The pure tree-based approaches do not benefit from convergence much, and also show a constant cost per iteration on the right part.
Since the Hybrid version switches to the successful Shallot approach after reaping the benefits of the tree,
it uses the least time overall.

\begin{figure}[tb!]\centering
    \includegraphics[width=.7\linewidth]{figures/legend.pdf}\\
    \begin{subfigure}{.48\linewidth}\centering
        \includegraphics[width=\linewidth]{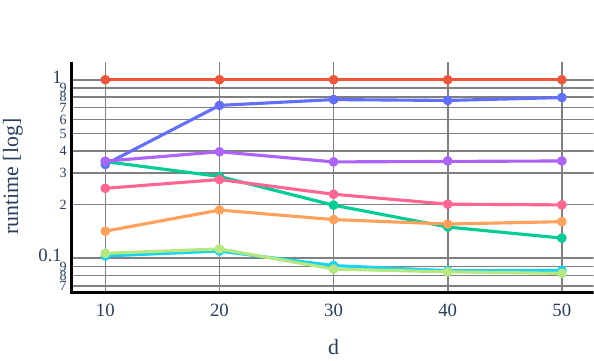}
        \caption{k = 400}
        \label{fig:d_scale}
    \end{subfigure}
    \hfill
    \begin{subfigure}{.48\linewidth}\centering
        \includegraphics[width=\linewidth]{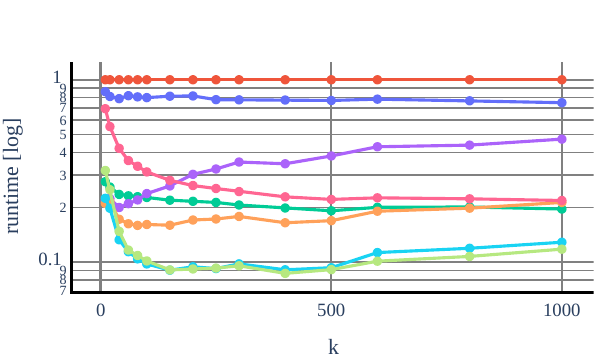}
        \caption{d = 30}
        \label{fig:k_scale}
    \end{subfigure}
    \caption{Runtime in relation to the Standard algorithm vs.{} $k$ respectively $d$ on the MNIST dataset}
    \label{fig:scaling}
\end{figure}

Next, we look at the MNIST dataset in \reffig{fig:scaling} to examine how the different algorithms scale with the dimensionality $d$ and the number of clusters $k$.
\reffig{fig:d_scale} shows the runtime in relation to the runtime of the Standard algorithm over multiple different dimensionalities.
Most algorithms except Elkan scale about the same as the Standard algorithm. 
Kanungo's algorithm only has a slight advantage over the standard algorithm and is not really well-suited for high dimensional data.
In high-dimensional data, the cost of distance computations increases, and hence the benefits here of
Elkan's algorithm over the alternatives become more pronounced, as also observed in other studies.

In \reffig{fig:k_scale}, we can observe that the scaling behavior with $k$ is slightly more interesting.
For very low $k$ the benefit of all acceleration techniques is rather low, especially for the tree accelerations, and Kanungo's algorithm does not improve for higher $k$.
Over all $k$, we can see that Hamerly's algorithm scales worse than the standard algorithm. Exponion and Shallot also show slightly worse scaling properties than Elkan and Cover-means. Here the improvement for our Hybrid approach increases, and switching strategies later could bring further advantages.

In \reftab{tab:dist} we extend our view to the other datasets, and give the number of distance calculations
relative to the Standard algorithm.
For the stored-bound algorithms, Hamerly always uses the most distance calculations,
followed by Exponion and Shallot.
Elkan's algorithm has the least of all, as is expected, except for the traffic dataset
that has a very high number of samples and is very beneficial for the tree-based methods that can assign many points at once.
Our Hybrid approach is usually slightly better than the Shallot algorithm but does not come near Elkan's $k$-means,
while the Cover-means based approach is usually worse than Exponion. We can observe that the performance
of the tree-based approaches (Kanungo's algorithm more so than the cover tree) appears to depend more on the dataset.
In particular, the low-dimensional Istanbul and traffic datasets are beneficial for the tree-based approaches,
whereas Kanungo's algorithm struggles with the complex KDD04 dataset and uses even more distance calculations than the Standard algorithm.

\begin{table}[bt!]
    \caption{Relative number of distance calculations compared to the Standard algorithm for $k{=}100$}
    \label{tab:dist}
    \begin{tabular}{lcccccccc}
    & CovType & Istanbul & KDD04 & Traffic & MNIST10 & MNIST30 & ALOI27 & ALOI64 \\
Kanungo    & 0.006 & \textbf{0.002} & 1.450 & \textbf{0.000} & 0.149 & 0.370 & 0.036 & 0.048 \\
Elkan       & \textbf{0.004} & \textbf{0.002} & \textbf{0.025} & 0.001 & \textbf{0.007} & \textbf{0.009} & \textbf{0.005} & \textbf{0.006} \\
Hamerly     & 0.099 & 0.078 & 0.364 & 0.090 & 0.198 & 0.213 & 0.229 & 0.253 \\
Exponion    & 0.016 & 0.010 & 0.341 & 0.009 & 0.075 & 0.130 & 0.060 & 0.075 \\
Shallot     & 0.012 & 0.006 & 0.311 & 0.006 & 0.034 & 0.061 & 0.030 & 0.043 \\
Cover-means  & 0.012 & 0.003 & 0.807 & 0.001 & 0.097 & 0.180 & 0.044 & 0.063 \\
Hybrid      & 0.005 & 0.003 & 0.310 & 0.003 & 0.031 & 0.057 & 0.027 & 0.038 \\
\end{tabular}
\end{table}
\begin{table}[bt!]
    \caption{Relative run time compared to the Standard algorithm for $k{=}100$}
    \label{tab:time100}
    \begin{tabular}{lcccccccc}
    & CovType & Istanbul & KDD04 & Traffic & MNIST10 & MNIST30 & ALOI27 & ALOI64 \\
Kanungo    & 0.068 & 0.123 & 4.035 & 0.182 & 0.470 & 0.798 & 0.133 & 0.130 \\
Elkan       & 0.114 & 0.520 & \textbf{0.193} & 0.652 & 0.454 & 0.226 & 0.180 & 0.104 \\
Hamerly     & 0.139 & 0.171 & 0.383 & 0.173 & 0.262 & 0.238 & 0.262 & 0.278 \\
Exponion    & 0.064 & 0.132 & 0.369 & 0.142 & 0.150 & 0.161 & 0.107 & 0.109 \\
Shallot     & 0.062 & 0.134 & 0.346 & 0.145 & \textbf{0.120} & \textbf{0.098} & 0.084 & 0.080 \\
Cover-means  & 0.072 & 0.092 & 1.121 & 0.135 & 0.352 & 0.313 & 0.138 & 0.123 \\
Hybrid      & \textbf{0.051} & \textbf{0.084} & 0.457 & \textbf{0.130} & 0.133 & 0.102 & \textbf{0.082} & \textbf{0.076} \\
\end{tabular}
\end{table}

As seen before, there may be overheads in the computations overlooked when counting only distance computations,
in particular for Elkan's algorithm.
\reftab{tab:time100} %
shows the total run time of all algorithms,
including the construction of the trees.
The KDD04 dataset shows that for high dimensional data, Elkan's algorithm often is the fastest
because saving distance computations is the most beneficial then.
While the additional bounds help for larger $k$ (as they are updated individually), the memory overhead becomes an issue quickly.
On the other datasets, the Shallot algorithm is usually the fastest state-of-the-art algorithm.
Our cover tree approach is most of the time faster than Kanungo's, Hamerly's and Elkan's algorithms, but it cannot compete with Exponion or Shallot.
Combining the cover tree with Shallot in our Hybrid approach leads to the overall best results. %
The results likely could be further improved by tuning the hyperparameters, e.g., changing from the Cover-means to the Shallot algorithm at some ``optimal'' point, or increasing the leaf size for the larger data sets, but we have not yet developed a heuristic for this and do not want to overfit to this benchmark.

\begin{table}[bt!]
    \caption{Relative runtime compared to the Standard algorithm with multiple restarts (parameter sweep to choose $k$)}
    \label{tab:overview}
    \begin{tabular}{lcccccccc}
& CovType & Istanbul & KDD04 & Traffic & MNIST10 & MNIST30 & ALOI27 & ALOI64 \\
Kanungo    & 0.040 & 0.112 & 5.090 & 0.162 & 0.409 & 0.903 & 0.114 & 0.116 \\
Elkan       & 0.093 & 0.609 & \textbf{0.171} & - & 0.351 & 0.187 & 0.121 & 0.065 \\
Hamerly     & 0.211 & 0.208 & 0.453 & 0.238 & 0.338 & 0.347 & 0.284 & 0.304 \\
Exponion    & 0.040 & 0.145 & 0.492 & 0.162 & 0.154 & 0.172 & 0.077 & 0.077 \\
Shallot     & 0.037 & 0.145 & 0.414 & 0.154 & \textbf{0.121} & 0.100 & 0.059 & 0.050 \\
Cover-means  & 0.028 & 0.059 & 1.015 & 0.093 & 0.272 & 0.248 & 0.086 & 0.077 \\
Hybrid      & \textbf{0.020} & \textbf{0.056} & 0.463 & \textbf{0.089} & 0.122 & \textbf{0.095} & \textbf{0.055} & \textbf{0.047} \\
\end{tabular}

\end{table}
Lastly, \reftab{tab:overview} shows the relative runtime for a complete parameter sweep for the individual datasets, as one would often need to do in practice when the ``true'' number of clusters $k$ is not known. Here, we measure the time it takes to run the algorithms for 10 different initializations (restarts, because $k$-means may converge to different local fix points) and 16 different values for $k$ (to find the best number of clusters). Then, the ``best'' clustering can be chosen by a heuristic such as the ``Elbow'' method, or any of the better alternatives~\cite{DBLP:journals/sigkdd/Schubert23}.
In this scenario, the dataset is used multiple times, and we can reuse the cover-tree we built to amortize the construction cost.
In this task, we see Elkan again be the fastest on difficult high dimensional data like the KDD04 dataset, but not being able to handle the complete traffic dataset because of memory overhead for storing all bounds.
On all other datasets, our Hybrid approach is the fastest except for the 10D MNIST dataset, where the Shallot algorithm is faster.

\section{Conclusion}
We show that tree-based $k$-means algorithms can be beneficial, in particular for huge datasets
because they can assign many points at once. Our new Cover-means approach outperforms the earlier
approaches based on the k-d tree on most datasets.
The use of the triangle inequality in cover tree $k$-means for pruning the set of candidate clusters
makes it easier to combine this with the other approaches than the bounding box-based approach of the k-d tree methods.
The cover tree also uses less memory, as storing the ball needs fewer parameters than storing a bounding box,
but also because the cover tree has a higher fan-out and lower depth than the k-d tree.
Our new method is in particular well-suited for clustering with a large number of clusters~$k$,
where scalability becomes more important.
For small~$k$, it will often be sufficient to cluster a subsample of the data~\cite{DBLP:conf/kdd/BradleyFR98} on a single CPU,
and we will usually find a sufficiently good result because of the stability properties of $k$-means~\cite{journals/annstat/Pollard81},
if a \emph{stable} result exists.

We also show how a hybrid of index-tree-based and stored-bound-based approaches combines the benefits of both
worlds and improves over the performance of state-of-the-art approaches for many scenarios.
While our hybrid approach that combines the cover tree with the Shallot algorithm is very basic,
there are new challenges when fully integrating both approaches in future work:
when the current hybrid switches to the Shallot strategy after a fixed number of iterations,
it no longer exploits redundancy in the dataset, but also uses individual bounds for each point.
The results on the large Traffic dataset suggest that we may want to be able to stick longer to
tree-based aggregation for performance for huge datasets, and we have not yet developed a heuristic for this.

The source code of this algorithm will be made available in the ELKI clustering toolkit~\cite{DBLP:conf/sisap/Schubert22},
and we hope that it lays the ground for future research on further accelerating $k$-means
by combining the strength of tree-based and stored-bounds-based algorithms.

\vfill\pagebreak
\bibliographystyle{splncs04}
\bibliography{literature}

\end{document}